\documentclass[format=manuscript]{acmart}
\usepackage{alltt}
\usepackage{bm}
\usepackage{listings}
\usepackage{hhline}
\usepackage{booktabs}
\usepackage{multirow}
\usepackage{subcaption}

\definecolor{background}{HTML}{F7F7F7}
\definecolor{keyword}{HTML}{37AC4A}
\definecolor{operator}{HTML}{A51DFF}
\definecolor{string}{HTML}{C03333}
\definecolor{comment}{HTML}{357979}

\lstdefinelanguage{python}{
  xleftmargin=4mm,
  numbers=left,
  numbersep=2mm,
  numberstyle=\tiny\color{gray},
  basicstyle=\fontsize{8}{9}\ttfamily,
  columns=fixed,
  basewidth=0.5em,
  stringstyle=\textcolor{string},
  showstringspaces=false,
  morestring=[b]",
  morestring=[b]""",
  morecomment=[l]\#,
  commentstyle=\textcolor{comment},
  morekeywords={and,as,assert,break,class,continue,def,del,elif,else,except,False,finally,for,from,global,if,import,in,is,lambda,None,nonlocal,not,or,pass,raise,return,True,try,while,with,yield},
  keywordstyle=\color{keyword}\bf\ttfamily,
  literate=
    *{>>}{{\bf\texttt{\color{operator}{>{}>}}}}{1}
    {\&}{{\bf\texttt{\color{operator}{\&}}}}{1}
    {|}{{\bf\texttt{\color{operator}{|}}}}{1}
    {=}{{\bf\texttt{\color{operator}{=}}}}{1}
    {(}{{\bf\texttt{\color{keyword}{(}}}}{1}
    {)}{{\bf\texttt{\color{keyword}{)}}}}{1}
    {[}{{\bf\texttt{\color{keyword}{[}}}}{1}
    {]}{{\bf\texttt{\color{keyword}{]}}}}{1}
    {\{}{{\bf\texttt{\color{keyword}{\char '173}}}}{1}
    {\}}{{\bf\texttt{\color{keyword}{\char '175}}}}{1},
}
\newcommand{\python}[1]{\lstinline[language=python]{#1}}
\newcommand*{\pyplain}[1]{\fontsize{8}{9}\texttt{#1}}

\newcommand*{\bfit}[1]{\textbf{\textit{#1}}}

\hypersetup{draft} 

\begin{document}

\title{An Empirical Study of Modular Bias Mitigators and Ensembles}

\author{Michael Feffer}
\affiliation{
  \institution{Carnegie Mellon University}
  \country{USA}
}

\author{Martin Hirzel}
\affiliation{
  \institution{IBM Research}
  \country{USA}
}

\author{Samuel C.~Hoffman}
\affiliation{
  \institution{IBM Research}
  \country{USA}
}

\author{Kiran Kate}
\affiliation{
  \institution{IBM Research}
  \country{USA}
}

\author{Parikshit Ram}
\affiliation{
  \institution{IBM Research}
  \country{USA}
}

\author{Avraham Shinnar}
\affiliation{
  \institution{IBM Research}
  \country{USA}
}

\begin{abstract}
There are several bias mitigators that can reduce algorithmic bias in
machine learning models but, unfortunately, the effect of mitigators on fairness is often not
stable when measured across different data splits.
A popular approach to train more stable models is ensemble learning.
Ensembles, such as bagging, boosting, voting, or stacking, have
been successful at making predictive performance more stable.
One might therefore ask whether we
can combine the advantages of bias mitigators and ensembles?
To explore this question, we first need bias mitigators and
ensembles to work together.
We built an open-source library enabling the modular composition of 10~mitigators,
4~ensembles, and their corresponding hyperparameters.
Based on this library, we empirically explored the space of
combinations on 13 datasets, including datasets commonly used in
fairness literature plus datasets newly curated by our library.
Furthermore, we distilled the results into a guidance diagram for
practitioners.
We hope this paper will contribute towards improving stability
in bias mitigation.

\end{abstract}

\maketitle

\section{Introduction}\label{sec:intro}

Algorithmic bias and discrimination in machine learning are a huge problem.
If learned estimators make biased predictions, they might discriminate
against underprivileged groups in various domains including job hiring,
healthcare, loan approvals, criminal justice, higher
education, and even child care.
These biased predictions can reduce diversity, for instance, in the workforce
of a company or in the student population of an educational institution.
Such lack of diversity can cause adverse business or educational outcomes.
In addition, several of the above-mentioned domains are governed by
laws and regulations that prohibit biased decisions.
And finally, biased decisions can severely damage the reputation of the
organization that makes them.
Of course, bias in machine learning is a sociotechnical problem that
cannot be solved with technical solutions alone.
That said, to make tangible progress, this paper focuses on
\emph{bias mitigators} that can reduce bias in machine learning models.
We acknowledge that bias mitigators can, at most, be a part of a larger
solution.

A bias mitigator either improves or replaces an existing
machine learning estimator (e.g., a classifier) so it makes less
biased predictions (e.g.,~class labels)
as measured by a fairness metric (e.g., disparate impact).
Unfortunately, bias mitigation often suffers from high volatility.
There is usually less training data available for underrepresented
groups.
Less data means the learned estimator has fewer examples to generalize
from for these groups. That in turn means the estimator is less stable
with respect to group fairness metrics, which are computed based on
predictive performance for different groups.
For instance, empirical studies (e.g.~\cite{bellamy_et_al_2018}) have
shown that volatility in fairness metrics tends to exceed
volatility in accuracy metrics.
With an unlucky train-test split, in the worst case, this volatility
can even cause a model to appear fair when measured on training data
while being unfair on production data.

Given that ensembles (e.g., bagging or boosting) can improve stability
for accuracy metrics, intuitively, we would expect that they can also
improve stability for group fairness metrics.
For instance, bagging ensembles work best when the base model is
unstable~\cite{witten_et_al_2016}: they turn instability from a
drawback into an advantage.
Prior work either explores bias mitigation without any
consideration of ensembles, or entangles the
two~\cite{kamiran_karim_zhang_2012,grgichlaca_et_al_2017,bhaskaruni_hu_lan_2019,kenfack_et_al_2021,mishler_kennedy_2021}.
In contrast, our paper hypothesizes that bias mitigators
and ensembles can be modular building blocks: instead of being
entangled with each other, they can be combined as needed.
The advantage of keeping ensembles and bias mitigators modular is
a larger space of possible combinations to explore.
Furthermore, when there are future advances in either ensembling or
bias mitigation, modularity helps extend these advances to their combination.

This paper explores the question,
``\emph{Can modular ensembles help with fairness, and if yes, how?}''
We conducted a comprehensive empirical study with
10~bias mitigators from AIF360~\cite{bellamy_et_al_2018};
bagging, boosting, voting, and stacking ensembles from the popular
scikit-learn library~\cite{buitinck_et_al_2013};
and 13~datasets of varying baseline fairness with sizes ranging from
\mbox{118 to 48,842} rows.
Our findings confirm the intuition that ensembles often improve
stability of not just accuracy but also the group fairness metrics
we explored.
Occasionally, ensembles even lead to better places in the combined
fairness/accuracy space.
However, the best configuration of mitigator and ensemble depends on
dataset characteristics, learning objectives, and even
worldviews~\cite{friedler_scheidegger_venkatasubramanian_2021}.
Therefore, this paper includes a guidance diagram that we
systematically distilled out of our extensive experimental results.

To support these experiments, we assembled a library of pluggable
ensembles, bias mitigators, and fairness datasets.
For the original components, we reused popular and well-established
open-source technologies including
scikit-learn~\cite{buitinck_et_al_2013},
pandas~\cite{mckinney_2011},
AIF360~\cite{bellamy_et_al_2018}, and
OpenML~\cite{vanschoren_et_al_2014}.
However, we found that out-of-the-box, these components were often not
able to plug-and-play with each other.
Hence, our library makes several new adaptations to get components to work
well together by exposing the right interfaces.
Since we wanted our library to be useful not just for research but
also for real-world adoption, we added thorough tests and
documentation and made everything available as open-source code
(\url{https://github.com/IBM/lale}).
Modular ensembles can have additional advantages related to fairness.
For instance, it has been shown that there is a fundamental trade-off between
fairness and accuracy~\cite{kamiran_calders_2012}.
Modular ensembles let data scientists navigate the fairness/accuracy
space by varying the mitigation of base estimators
in the ensemble or by mixing different kinds of
mitigators~\cite{kenfack_et_al_2021}.

To summarize, this paper makes the following contributions:

\begin{enumerate}
  \item An open-source library of modular ensembles and bias
    mitigators, c.f.\ Section~\ref{sec:library}.
  \item An empirical study of ensembles and bias
    mitigators, c.f.\ Section~\ref{sec:empirical}.
  \item A guidance diagram to help practitioners combine ensembles
    with bias mitigators, c.f.\ Section~\ref{sec:guidance}.
\end{enumerate}

Overall, this paper answers the question
``\emph{Can modular ensembles help with fairness?}''
with ``\emph{yes}''.
The follow-up question ``\emph{If yes, how?}'' is more important but
harder to address. This paper answers it
by showing that ensembles improve fairness stability, i.e., they yield
estimators whose fairness generalizes better to new data.
This is a step towards a future where we can better trust
machine learning to be fair.

\section{Related Work}\label{sec:related}

A few pieces of prior work have used ensembles for
fairness, but they use specialized ensembles and bias mitigators,
in contrast to our work, which uses off-the-shelf modular components.
The \emph{discrimination-aware ensemble} uses a heterogeneous
collection of base estimators~\cite{kamiran_karim_zhang_2012}. When
all base estimators agree, the ensemble returns the consensus
prediction, otherwise, it classifies instances as positive if and only
if they belong to the unprivileged group. This can be viewed as a form
of stacking ensemble with a simple policy-based final estimator.
The \emph{random ensemble} also uses a heterogeneous collection of
base estimators, and picks one of them at random to make a
prediction~\cite{grgichlaca_et_al_2017}. This can be viewed as a form
of stacking ensemble with a random final estimator. The paper offers a
synthetic case where the resulting ensemble is both more fair and more
accurate than all base estimators, but lacks experiments with real
datasets.
The \emph{exponentiated gradient reduction} trains a sequence of base
estimators using a game, where one player seeks to maximize fairness
violations by the estimators so far and the other player seeks to
build a fairer next estimator~\cite{agarwal_et_al_2018}.  In the end,
for predictions, it uses weights to pick a random base estimator.
Even though the authors do not frame their algorithm in ensemble
terminology, it has aspects reminiscent of boosting ensembles.
The \emph{fair AdaBoost} algorithm modifies boosting ensembles to
boost not for accuracy but for fairness~\cite{bhaskaruni_hu_lan_2019}.
It trains a sequence of base estimators, where the training data for
the next estimator puts more weight on instances that the previous
estimator predicted unfairly, based on an individual fairness
measure.  In the end, for predictions, it gives a base estimator
higher weight if it was fair on more instances from the
training set.
The \emph{fair voting ensemble} uses a heterogeneous collection of
base estimators~\cite{kenfack_et_al_2021}. For each prediction, it
votes among the base estimators $\phi_t$, \mbox{$t\in 0..n-1$}, with weights
\mbox{$W_t=\alpha\cdot A_t/(\Sigma_{t=0}^{n-1}A_j)+(1-\alpha)\cdot F_t/(\Sigma_{t=0}^{n-1}F_j)$},
where $A_t$ is an accuracy metric and $F_t$ is a fairness metric.
The \emph{fair double ensemble} algorithm uses stacked predictors,
where the final estimator is linear, with a novel approach to train
the weights of the final estimator to satisfy a system of accuracy and
fairness constraints~\cite{mishler_kennedy_2021}.

Each of the above-listed approaches uses an ensemble-specific bias mitigator,
whereas we experiment with ten different off-the-shelf modular
mitigators.  Similarly, each of these approaches uses one specific
kind of ensemble, whereas we experiment with off-the-shelf modular
implementations of bagging, boosting, voting, and stacking.
Using off-the-shelf mitigators and ensembles facilitates plug-and-play
between the best available independently-developed implementations.
Given that fair learning is a rapidly evolving field, specialized mitigator-ensemble
combinations may be appropriate. However, we believe that it is still useful to
study off-the-shelf tools given that these have established open-source
implementations and are more readily available, as opposed to specialized
tools that are still in the process of being hardened.
Out of the work on fairness with ensembles discussed above,
one paper has an experimental
evaluation with five datasets~\cite{agarwal_et_al_2018} and the other
papers use at most three datasets. In contrast, we use
13~datasets. Finally, unlike these earlier papers, our paper
specifically explores fairness stability, extracting that as one of
the goals for our auto-generated guidance diagram.

Our work takes inspiration from earlier empirical studies and
comparisons of fairness techniques
\cite{biswas_rajan_2021,friedler_et_al_2019,holstein_et_al_2019,lee_singh_2021,singh_et_al_2021,valentim_lourencco_antunes_2019,yang_et_al_2020},
which help practitioners and researchers better understand the state
of the art. But unlike these works, we experiment with ensembles and
with fairness stability.

Our work offers a new library of bias mitigators.
While there have been excellent prior fairness toolkits such as
ThemisML~\cite{bantilan_2017}, AIF360~\cite{bellamy_et_al_2018}, and
FairLearn~\cite{agarwal_et_al_2018}, none of them support ensembles.
Ours is the first that is modular enough to investigate a large space of
unexplored mitigator-ensemble combinations.
%
We previously published some aspects of our library in a non-archival
workshop with no official proceedings, but that paper did not yet
discuss ensembles~\cite{hirzel_kate_ram_2021}.

\section{Library and Datasets}\label{sec:library}
Our experiments were made possible by multiple Python libraries and 13 different
datasets. The subsections that follow provide more information about these libraries and datasets
and describe how they interact with each other.

\subsection{Lale and fairness metrics}\label{sec:lale}
Lale is an open-source library for semi-automated data science~\cite{baudart_et_al_2021}.
It automates parts of the iterative model-building process and serves as an intuitive frontend for
scikit-learn~\cite{buitinck_et_al_2013} and several other machine learning libraries.
Aside from our experiments, one contribution of our work is implementing Lale compatibility
with another library: the \mbox{AI Fairness 360} (AIF360) Toolkit~\cite{bellamy_et_al_2018},
especially with regard to interoperability between scikit-learn's ensemble learning algorithms and AIF360's bias mitigation algorithms
through Lale's operator framework. Building this functionality and handling quirks in
AIF360 and scikit-learn as necessary was a key part of this research.

As Lale is compatible with scikit-learn, its models take \pyplain{X}
and \pyplain{y} arguments corresponding to features and labels,
respectively.
To provide a unified API to fairness metrics and mitigators, we added
a format to Lale for specifying information about favorable and
unfavorable labels as well as privileged and unprivileged groups that
reflect bias in a dataset.
This is done via a \pyplain{fairness\_info} dictionary with fields for
\pyplain{favorable\_labels} and \pyplain{protected\_attributes}
that represent favorable outcomes and privileged groups.
We then wrote wrappers for bias mitigators (e.g.,
\pyplain{DisparateImpactRemover}) and metrics (e.g.,
\pyplain{disparate\_impact}) in AIF360 that understand this
\pyplain{fairness\_info} format in addition to the usual scikit-learn
style \pyplain{X} and \pyplain{y} arguments, given as pandas
dataframes.
Fig.~\ref{lst:lale-mitigator-fit-example} shows an example.

\begin{figure}[!h]
\begin{lstlisting}[language=python]
fairness_info = {
    "favorable_labels": [1],  # values of `y` that indicate a favorable outcome
    "protected_attributes": [  # columns of `X` and values that indicate a privileged group
        {"feature": "race", "reference_group": ["White"]},
        {"feature": "sex", "reference_group": ["male div/sep", "male mar/wid", "male single"]},
    ],
}
mitigator = DisparateImpactRemover(**fairness_info)
pipeline = make_pipeline(mitigator, DecisionTreeClassifier())
trained = pipeline.fit(X, y)
predictions = trained.predict(test_X)
\end{lstlisting}
\caption{Sample code showing bias mitigation workflow with Lale and AIF360 through fairness\_info.}
\label{lst:lale-mitigator-fit-example}
\end{figure}

While the example in Fig.~\ref{lst:lale-mitigator-fit-example}
configures operators with their default hyperparameters, the
programming model also supports more general configuration. For instance,
\mbox{\python{DisparateImpactRemover(**fairness_info, repair_level=0.8)}}
tunes the mitigator, and
\mbox{\python{DecisionTreeClassifier(max_depth=10, criterion="entropy")}}
tunes the estimator.
Once models are trained, Lale assists in auditing their performance from
both accuracy and fairness standpoints. Metrics used here include, but are not limited to:
\begin{itemize}
  \item Accuracy: ratio of number of examples correctly predicted to total number of examples predicted
  \item $F_1$ score: harmonic mean of precision and recall
  \item Disparate Impact: ratio of positive outcomes for unprivileged group to positive outcomes for privileged group (as described in~\cite{feldman_et_al_2015})
\end{itemize}

For convenience, just as scikit-learn provides \pyplain{scorer} objects for
accuracy metrics, we added Lale \pyplain{scorer} objects for fairness metrics.
Fig.~\ref{lst:lale-scorer-example} shows an example demonstrating
how to use scorers from both packages.

\begin{figure}[!h]
\begin{lstlisting}[language=python, firstnumber=12]
accuracy_scorer = sklearn.metrics.make_scorer(sklearn.metrics.accuracy_score)
accuracy_measured = accuracy_scorer(trained, test_X, test_y)
di_scorer = lale.lib.aif360.disparate_impact(**fairness_info)  # uses fairness_info defined in Line 1
di_measured = di_scorer(trained, test_X, test_y)
\end{lstlisting}
\caption{Sample code showing scorer objects.}
\label{lst:lale-scorer-example}
\end{figure}

Two types of the most commonly used fairness metrics are
group fairness metrics (like disparate impact) and individual fairness metrics
(like those described in~\cite{dwork2012fairness} that "treat similar individuals similarly").
Since the mitigators in our experiments focus on group fairness,
our experiments focus on group fairness metrics.

\subsection{Ensembles}\label{sec:ensembles}
The main idea behind ensemble learning is to use multiple weak models to form one strong model.
This can be done by training more models on data that is difficult to fit, combining predictions of models
trained on various subsets of the input data, or combining predictions of different types of models to
improve robustness through model diversity~\cite{witten_et_al_2016}.
Scikit-learn supports several types of ensembles~\cite{buitinck_et_al_2013}.
We use four in our experiments, specifically classifier implementations
from scikit-learn that are supported by Lale.
These are summarized in Table~\ref{tab:ensemble-types}.

\begin{table}
  \begin{tabular}{lp{0.7\textwidth}c}
    \toprule
    \textbf{Ensemble} & \textbf{Algorithm} & \textbf{Composition}\\
    \midrule
    Bagging & Train $n$ base estimators in parallel on random subsets of training data. & Homogeneous\\
    \addlinespace[0.5em]
    Boosting & Train $n$ base estimators in series where each subsequent base estimator is fit on data incorrectly classified by a previous base estimator. & Homogeneous\\
    \addlinespace[0.5em]
    Voting & Train $n$ base estimators in parallel and determine overall predictions by choosing the most frequently occurring output from base estimators. & Heterogeneous\\
    \addlinespace[0.5em]
    Stacking & Train $n$ base estimators in parallel as well as a \textit{final estimator} that makes overall predictions given outputs of the other $n$ base estimators as input. In addition, the final estimator can optionally also use the original input data (\pyplain{passthrough=True}). & Heterogeneous\\
    \bottomrule
  \end{tabular}
  \caption{Overview of ensemble types used in our experiments.}
  \label{tab:ensemble-types}
\end{table}

Following scikit-learn, we use the following terminology to characterize ensembles:
A \emph{base estimator} is an estimator that serves as a building
block for the ensemble.
An ensemble supports one of two \emph{composition} types: whether the
ensemble consists of identical base estimators (\emph{homogeneous}) or
can consist of different ones (\emph{heterogeneous}).
Similarly, each ensemble supports one of two \emph{training styles}:
whether the ensemble trains base estimators one at a time sequentially
(\emph{series}) or independently from each other (\emph{parallel}).

Additionally, it is necessary to choose specific base estimators to use in the
ensembles. For the experiments in this paper,
this choice was constrained by the fact that both boosting ensembles
and post-estimator bias mitigators require base estimators that
can return not just target labels but class probabilities
(i.e., \pyplain{predict\_proba} in scikit-learn).
While other ensembles do not impose that restriction, they can still
benefit from \pyplain{predict\_proba} if it is present, such as stacking.
Furthermore, using the same base estimators across all experiments
helps in making apples-to-apples comparisons between configurations.
Specifically, for the homogeneous ensembles (bagging and boosting), we
used their most common base estimator in practice: the
decision-tree classifier.
For the heterogeneous ensembles (voting and stacking), we used a set
of base estimators that are typical in common practice: XGBoost~\cite{chen_guestrin_2016}, random
forest, k-nearest neighbors, and support vector machines.
Finally, for stacking, we also used XGBoost as the final estimator.

\subsection{Mitigators}\label{sec:mitigators}

We added support in Lale for bias mitigation from AIF360~\cite{bellamy_et_al_2018}.
AIF360 distinguishes three kinds of mitigators for improving group fairness:
\emph{pre-estimator mitigators}, which are learned input manipulations
that reduce bias in the data sent to downstream estimators;
\emph{in-estimator mitigators}, which are specialized estimators that
directly incorporate debiasing into their training; and
\emph{post-estimator mitigators}, which attempt to reduce bias in
predictions made by an upstream estimator.
Table \ref{tab:mitigators} lists the specific mitigators along with
their hyperparameters and originating papers.

\begin{table}
  \begin{tabular}{cl@{\hspace{-2em}}rp{0.35\textwidth}}
    \toprule
    \textbf{Kind} & \textbf{Mitigator} & \multicolumn{2}{c}{\textbf{Hyperparameters}}\\
    \cmidrule(lr){3-4}
                       &               & \textbf{Name}  & \textbf{Description} \\
    \midrule
    \multirow{6}{*}{\rotatebox[origin=c]{90}{pre-estimator}}
    & DisparateImpactRemover \cite{feldman_et_al_2015} & \bfit{repair\_level} & repair amount \\
    \addlinespace[0.3em]
        & LFR \cite{zemel_et_al_2013} & $k$ & number of prototypes \\
        &               & $A_x$ & input reconstruction quality term weight \\
        &               & $A_y$ & output prediction error term weight \\
        &               & $\bfit{A}_\bfit{z}$ & fairness constraint term weight \\
    \addlinespace[0.3em]
    & Reweighing \cite{kamiran_calders_2012} & N/A &  \\
    \midrule
    \multirow{13}{*}{\rotatebox[origin=c]{90}{in-estimator}}
    & AdversarialDebiasing \cite{zhang_lemoine_mitchell_2018} & \bfit{adversary\_loss\_weight} & strength of adversarial loss \\
                         &              & \textit{num\_epochs} & number of training epochs \\
                         &              & \textit{batch\_size} & batch size \\
                         &              & \textit{classifier\_num\_hidden\_units} & number of hidden units \\
                         &              & \textit{debias} & learn classifier with or without debiasing \\
    \addlinespace[0.3em]
    & GerryFairClassifier \cite{kearns_et_al_2018} & $C$ & maximum L1 Norm for the dual variables \\
                        &              & \textit{max\_iters} & time horizon for fictitious play dynamic \\
                        &              & $\boldsymbol\gamma$ & fairness approximation parameter \\
                        &              & \textit{fairness\_def} & fairness notion \\
                        &              & \textit{predictor} & hypothesis class for the Learner \\
    \addlinespace[0.3em]
    & MetaFairClassifier \cite{celis_et_al_2019} & $\boldsymbol\tau$ & fairness penalty parameter \\
                       &              & \textit{type} & type of fairness metric to be used \\
    \addlinespace[0.3em]
    & PrejudiceRemover \cite{kamishima_et_al_2012} & $\boldsymbol\eta$ & fairness penalty parameter \\
    \midrule
    \multirow{3}{*}{\rotatebox[origin=c]{90}{post-estim.}}
    & CalibratedEqOddsPostprocessing \cite{pleiss_et_al_2017}    & \textit{cost\_constraint} & fpr, fnr, or weighted\\
    \addlinespace[0.4em]
    & EqOddsPostprocessing \cite{hardt_et_al_2016}               & \multicolumn{2}{c}{(not used in experiments due to lack of \pyplain{predict\_proba})}\\
    \addlinespace[0.4em]
    & RejectOptionClassification \cite{kamiran_karim_zhang_2012} & \multicolumn{2}{c}{(not used in experiments due to lack of \pyplain{predict\_proba})}\\
    \bottomrule
  \end{tabular}
  \caption{Mitigators and their hyperparameters and originating papers. Hyperparameter descriptions from AIF360 documentation.
  Bolded hyperparameters control mitigation strength.
  All mitigators support \textit{favorable\_labels} and
  \textit{protected\_attributes} from Section~\ref{sec:lale}.}
  \label{tab:mitigators}
\end{table}

\begin{figure}
  \centering
  \begin{tabular}{c@{\hskip0.5em}l@{\hskip-0.5em}l@{\hskip1em}l@{\hskip1em}l@{\hskip-0.5em}l}
    \toprule
     & \multicolumn{2}{c}{\textbf{pre-estimator}} & \multicolumn{1}{c}{\textbf{in-estimator}} & \multicolumn{2}{c}{\textbf{post-estimator}} \\
    \cmidrule(lr){2-3}\cmidrule(lr){4-4}\cmidrule(lr){5-6}
    & \hskip0.5em\textbf{estimator-level} & \hskip0.5em\textbf{ensemble-level} & \hskip0.5em\textbf{estimator-level} & \hskip0.5em\textbf{estimator-level} & \hskip0.5em\textbf{ensemble-level} \\
    \midrule
    \addlinespace[0.5em]
    \multirow{1}{*}[2.5em]{\rotatebox[origin=c]{90}{\textbf{bagging}}} & \includegraphics[scale=0.6]{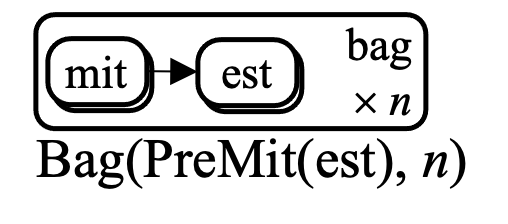} & \includegraphics[scale=0.6]{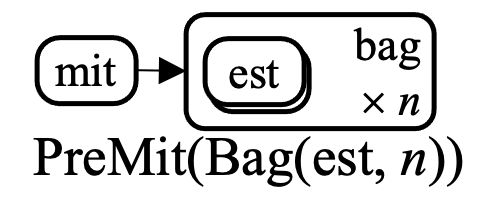} & \includegraphics[scale=0.6]{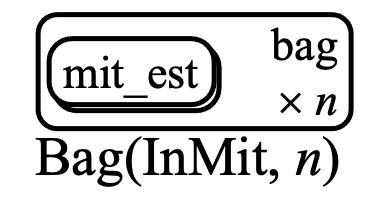} & \includegraphics[scale=0.6]{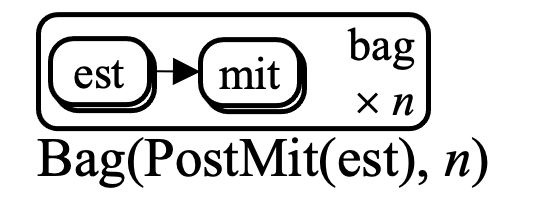} & \includegraphics[scale=0.6]{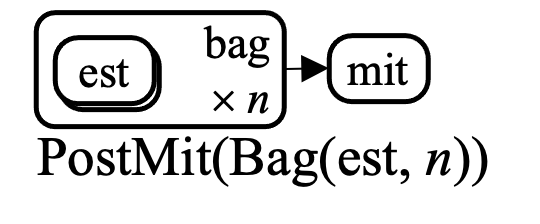} \\
    \addlinespace[0.5em]
    \multirow{1}{*}[2.5em]{\rotatebox[origin=c]{90}{\textbf{boosting}}} & \includegraphics[scale=0.6]{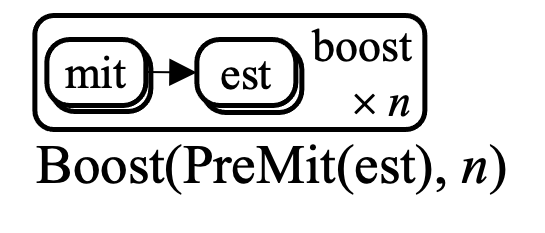} & \includegraphics[scale=0.6]{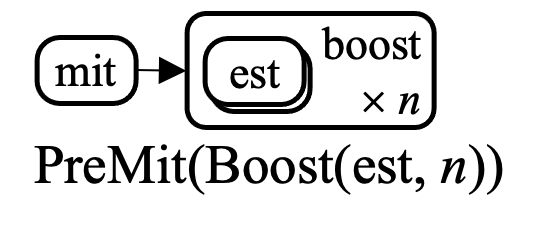} & \includegraphics[scale=0.6]{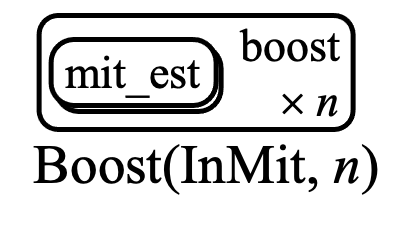} & \includegraphics[scale=0.6]{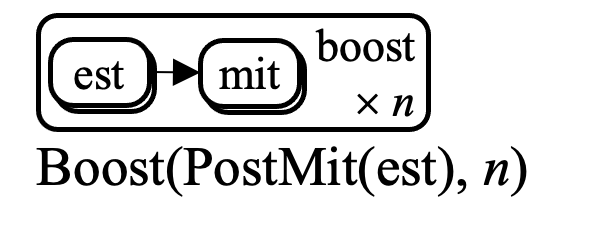} & \includegraphics[scale=0.6]{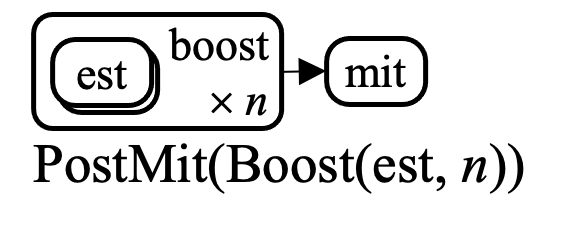} \\
    \addlinespace[0.5em]
    \multirow{1}{*}[3.5em]{\rotatebox[origin=c]{90}{\textbf{voting}}} & \includegraphics[scale=0.6]{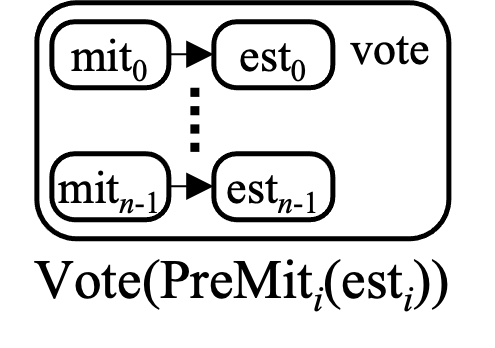} & \includegraphics[scale=0.6]{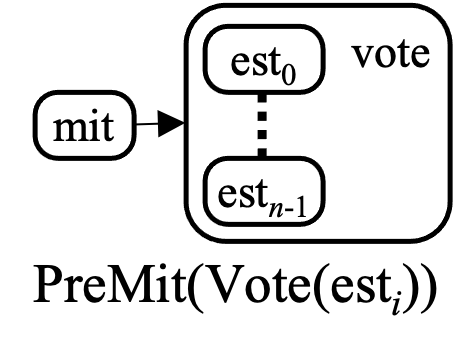} & \includegraphics[scale=0.6]{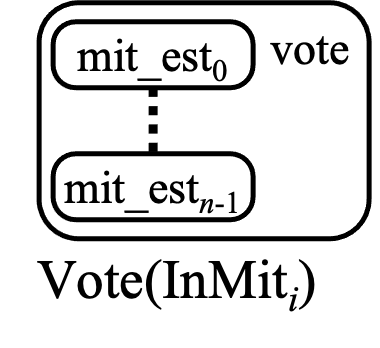} & \includegraphics[scale=0.6]{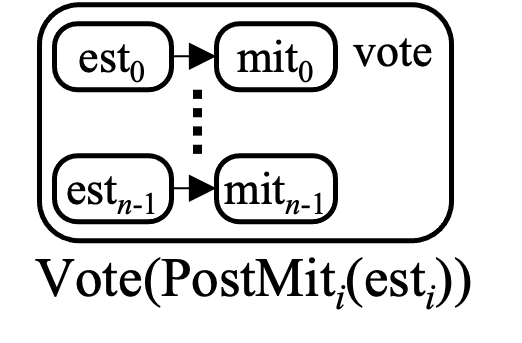} & \includegraphics[scale=0.6]{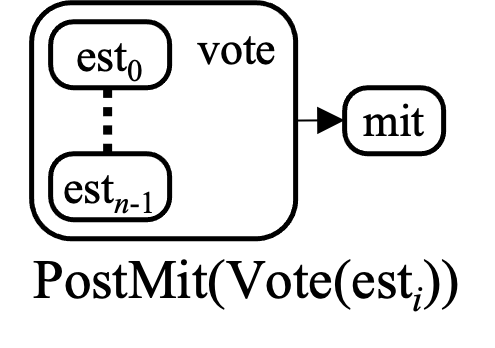} \\
    \addlinespace[0.5em]
    \multirow{2}{*}[2em]{\rotatebox[origin=c]{90}{\textbf{stacking}}} & \includegraphics[scale=0.6]{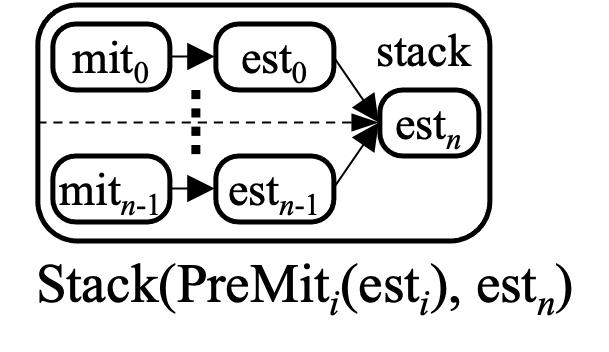} & \includegraphics[scale=0.6]{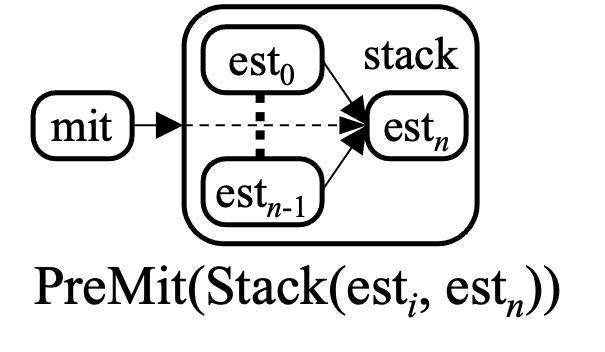} & \includegraphics[scale=0.6]{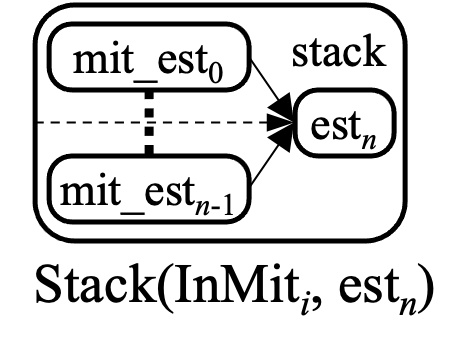} & \includegraphics[scale=0.6]{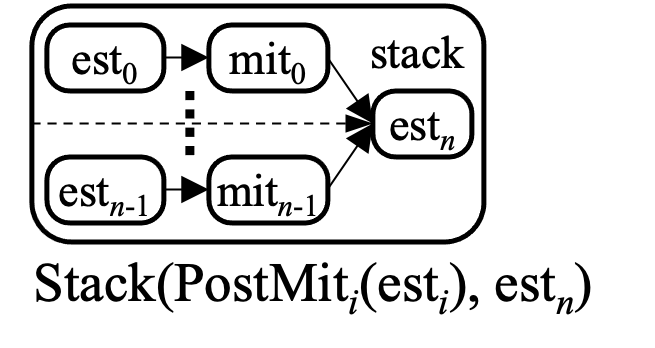} & \includegraphics[scale=0.6]{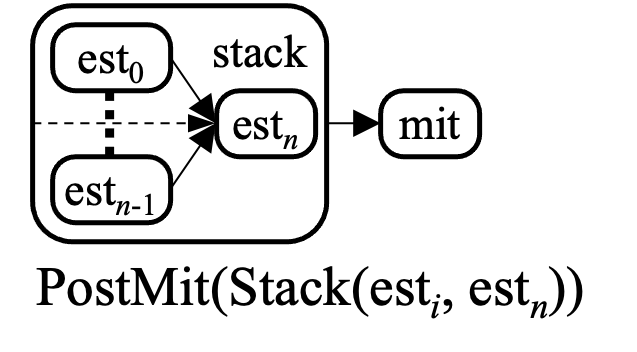} \\
     & \includegraphics[scale=0.6]{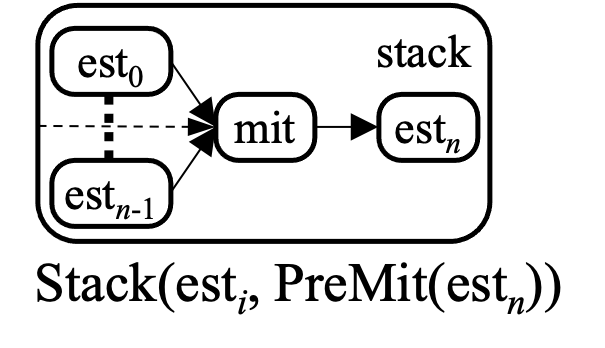} & & \includegraphics[scale=0.6]{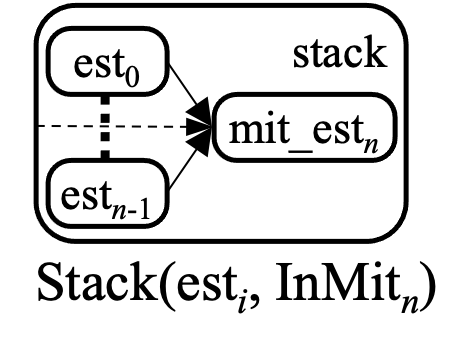} & \includegraphics[scale=0.6]{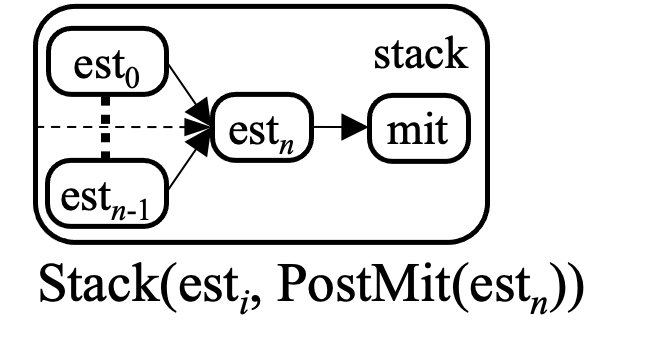} & \\
    \bottomrule
  \end{tabular}
  \caption{\label{fig:standard_model}Combinations of ensembles and mitigators.
    For stacking, the \textit{passthrough} option is represented by a
    dashed horizontal arrow.}
\vspace*{3mm}
\begin{lstlisting}[language=python]
StackingClassifier(
    estimators=[
        XGBClassifier(use_label_encoder=False, verbosity=0),
        RandomForestClassifier(),
        KNeighborsClassifier(),
        SVC(probability=True),
    ],
    final_estimator=make_pipeline(
        LFR(**fairness_info, k=5, Ax=0.01, Ay=10, Az=5),
        XGBClassifier(use_label_encoder=False, verbosity=0)
    ),
    passthrough=False
)
\end{lstlisting}
\caption{\label{lst:lale-mitigator-ensemble-example}Possible instantiation for the pseudo-code \pyplain{Stack(est$_i$, PreMit(est$_n$))} using actual Python code in our library.}
\end{figure}

Fig.~\ref{fig:standard_model} visualizes the combinations of
ensemble types and mitigator kinds we explored in our experiments.
It also shows each combination as pseudo-code, using the following notation.
\pyplain{PreMit(est)} applies a pre-estimator mitigator before an
estimator~\pyplain{est};
\pyplain{InMit} denotes an in-estimator mitigator, which is itself an
estimator;
and \pyplain{PostMit(est)} applies a post-estimator mitigator after an
estimator~\pyplain{est}.
\pyplain{Bag(est, n)} is short for \pyplain{BaggingClassifier} with
\pyplain{n} instances of base estimator \pyplain{est};
\pyplain{Boost(est, n)} is short for \pyplain{AdaBoostClassifier} with
\pyplain{n} instances of base estimator \pyplain{est};
\pyplain{Vote(est$_i$)} applies a \pyplain{VotingClassifier} to a list
of base estimators \pyplain{est$_i$};
and \pyplain{Stack(est$_i$, est$_n$)} applies a
\pyplain{StackingClassifier} to a list of base estimators
\pyplain{est$_i$} and a final estimator \pyplain{est$_n$}.
The pseudo-code notation is a short-hand for the actual code one can write
with our library, shown in Fig.~\ref{lst:lale-mitigator-ensemble-example}.
Fig.~\ref{fig:standard_model} highlights the modularity of our approach.
Mitigation strategies can be applied at the level of either the base
estimator or the entire ensemble.
However, it turns out that by the fundamental nature of some ensembles
and mitigators, not all combinations are feasible.
We had to limit ourselves to less than the full Cartesian product for
the following reasons.

First, post-estimator mitigators typically do not implement a \pyplain{predict\_proba} function as described in the previous subsection.
This functionality is required for some ensemble methods and recommended for others. To that end,
in the process of ensuring that AIF360 mitigators could work with scikit-learn ensembles, we ended up exposing
\pyplain{predict\_proba} functionality not exposed by AIF360 by default but produced by underlying
in-estimator and post-estimator mitigators. While we were able to get \pyplain{predict\_proba} working
for all of the in-estimator mitigators we wanted to test,
calibrating probabilities from post-estimator mitigators has been
shown to be tricky~\cite{pleiss_et_al_2017}. Hence, we only exposed
it for \pyplain{CalibratedEqOddsPostprocessing}.

Additionally, it is impossible to apply an in-estimator mitigator at the ensemble level
because in that case, both the ensemble and mitigator would be performing the same task --- estimation --- so there is no way to combine them.
Results for that part of the product are undefined and excluded from our analysis.
Finally, we decided to omit some combinations that are technically
feasible but less interesting to explore.
For example, it is possible to mitigate at multiple points, say, at both the
ensemble and estimator level of bagging or both the base and final estimators of
a stacking ensemble.
While our library supports these configurations, we elided them
from Fig.~\ref{fig:standard_model} and from our experiments.

\subsection{Datasets}\label{sec:dataset}

\begin{table}[]
  \begin{tabular}{lp{0.35\textwidth}lrrr}
  \toprule
  \textbf{Dataset}        & \textbf{Description}                             & \textbf{Privileged group(s)} & $\bm{N_\textit{rows}}$ & $\bm{N_\textit{cols}}$ & \textbf{DI}\\
  \midrule
  COMPAS Violent & ProPublica data from audit of Northpointe's recidivism algorithm but only considering violent recidivism       & White women & 3,377    & 10       & 0.822       \\
  \addlinespace[0.3em]
  Credit-g       & German Credit dataset quantifying credit risk  & Men and older people                                          & 1,000    & 58       & 0.748        \\
  \addlinespace[0.3em]
  COMPAS         & ProPublica data from audit of Northpointe's recidivism algorithm  &   White women                                     & 5,278    & 10       & 0.687        \\
  \addlinespace[0.3em]
  Ricci          & Test scores from fire department promotion exam with demographic info and promotion result & White men             & 118      & 6        & 0.498        \\
  \addlinespace[0.3em]
  TAE            & Teacher Assistant Evaluation results from U Wisconsin, Madison & Native English speakers                         & 151      & 6        & 0.449        \\
  \addlinespace[0.3em]
  Titanic        & Demographic info of Titanic passengers and whether they survived & Women and children                               & 1,309    & 37       & 0.263        \\
  \midrule
  SpeedDating    & Preferences of participants in experimental speed dating events at Columbia Business School  & Same race            & 8,378    & 70       & 0.853       \\
  \addlinespace[0.3em]
  Bank           & Data from Portuguese bank marketing campaign predicting whether client will subscribe to a term deposit & Older people  & 45,211   & 51       & 0.840        \\
  \addlinespace[0.3em]
  MEPS 19        & Utilization results from Panel 19 of Medical Expenditure Panel Survey &      White individuals         & 15,830   & 138      & 0.490        \\
  \addlinespace[0.3em]
  MEPS 20        & Same as MEPS 19 except for Panel 20          &  White individuals & 17,570   & 138      & 0.488        \\
  \addlinespace[0.3em]
  Nursery        & Nursery school application results during a competitive time period in Ljubljana, Slovenia & ``Pretentious parents'' & 12,960   & 25       & 0.461        \\
  \addlinespace[0.3em]
  MEPS 21        & Same as MEPS 19 except for Panel 21          &  White individuals & 15,675   & 138      & 0.451        \\
  \addlinespace[0.3em]
  Adult          & 1994 US Census data predicting salary over \$50K &      White men                                 & 48,842   & 100      & 0.277        \\
  \bottomrule
  \end{tabular}
  \caption{Qualitative and quantitative summary information of the datasets. The datasets are ordered by first partitioning by whether they contain at least 8,000 rows
  (we picked 8,000 to get a roughly even split; the partition is represented by the horizontal line in the middle of the table) and then sorting by descending baseline disparate impact (DI).
  Values for the number of rows ($N_\textit{rows}$), number of columns ($N_\textit{cols}$), and baseline disparate impact displayed here are computed \emph{after} preprocessing techniques are applied.}
  \label{tab:dataset-table}
\end{table}

We gathered the datasets for our experiments from OpenML~\cite{vanschoren_et_al_2014}. Some of these datasets
have been used extensively as benchmarks in other parts of the algorithmic fairness literature (including but not limited
to COMPAS, Adult, and Credit-g). We pulled novel datasets from OpenML based on whether they had demographic data that
could be considered protected attributes (such as race, age, or gender) and there were associated baseline levels of disparate impact
found in the dataset. This dataset discovery process yielded additional datasets like TAE, Titanic, and SpeedDating.

In all, we used 13 datasets in our research, summarized in Table \ref{tab:dataset-table}. When running experiments,
we split the datasets using stratification by not just the target
labels but also the protected attributes~\cite{hirzel_kate_ram_2021}.
This stratification approach leads to moderately more homogeneous
fairness results across different splits. We
use \pyplain{pandas} for data cleaning and preprocessing operations~\cite{mckinney_2011}. The exact details of the preprocessing
operations we use can be found in the open-source code for our library for reproducibility, but at a high level, the preprocessing
generally involves operations such as:
\begin{itemize}
  \item Removing columns that are irrelevant to model fitting (e.g., in the case of Titanic, such columns include ones with names, ticket type, cabin, and final destination).
  \item Binarizing protected attribute and outcome values and condensing them to one column each.
  \item Producing \texttt{fairness\_info} based on these operations.
  \item Standardizing (subtracting the mean and dividing by the standard deviation) each numerical column. The same coefficients (mean and standard deviation) learned at training time are applied at test time.
\end{itemize}

\section{Empirical Study}\label{sec:empirical}

This section uses our library of bias mitigators, ensembles, and
datasets to empirically explore their combinations.

\subsection{Methods}\label{sec:methods}

To make sense of our experimental results, and to ask and answer
research questions about fairness and ensembles, we first had to
narrow our experiments to a reasonably-sized space.
The entire space of all possible combinations is large due to the
combinatorial effect of the sets of possible choices:
collectively, the ten bias mitigators have many
hyperparameters (see Table~\ref{tab:mitigators});
the four ensembles also have hyperparameters of their own (e.g.,
$n$ and \textit{passthrough} in Fig.~\ref{fig:standard_model});
and we are experimenting with thirteen datasets (see
Table~\ref{tab:dataset-table}).
On top of that, we are exploring five different performance dimensions:
we measure predictive performance as F1 score, precision, and recall;
we measure fairness performance as disparate impact, equal opportunity difference, statistical parity difference, and average odds difference;
we quantify fairness volatility based on the standard deviation of the aforementioned fairness metrics;
we measure time efficiency in seconds;
and we measure memory efficiency in megabytes.

Therefore, we organize our experiments into two steps.
The first step is a preliminary search that finds ``best'' mitigators
without ensembles.
Since mitigators without ensembles have been studied elsewhere, this
paper does not present detailed results for this step of the
experiments.
Instead, this paper focuses on the second step, which is the main set
of experiments with ensembles, described in Section~\ref{sec:results}.
The second step uses only the mitigator configurations selected by
the first step.
By limiting the second step to fewer mitigator configurations,
we can more easily attribute performance differences to changes in
ensembling configurations.

In the first step, the main difficulty is how to decide what
configured mitigators (see Table~\ref{tab:mitigators}) are ``best''.
Since we are doing an empirical study, we mean \emph{best} in an
empirical sense of best encountered and picked during the search,
not in a theoretical sense of optimality.
That said, we still need to define what to consider best given
the different dimensions of performance, mitigators, datasets, etc.
To this end, we first run separate grid searches for each dataset,
exploring bias mitigators with their hyperparameters.
We run each configuration with five trials of 3-fold cross validation,
where splits are stratified not just by outcome labels but also by
protected groups~\cite{hirzel_kate_ram_2021}.
We group the grid search results by dataset and mitigator \emph{kind}:
for each dataset, we consider three sets of mitigator configurations,
one each for pre-, in-, and post-estimator mitigation.
More specifically, for pre-estimator mitigation, the group contains
three mitigators and their hyperparameters; for in-estimator
mitigation, the group contains four mitigators and their
hyperparameters; and for post-estimator mitigation, the group
contains only one mitigator, \pyplain{CalibratedEqOddsPostprocessing},
and its hyperparameters, as it is the only mitigator with
\pyplain{predict\_proba} in that category.

Given the results for each of the 39~groups (3~mitigator kinds
$\times$ 13~datasets), the first step then needs to pick a best
configuration in each group.
Picking a best configuration is complicated by the five often
conflicting performance dimensions.
The relative priorities between the performance dimensions depend on
the usage scenario.
After data exploration and discussion, we settled on the following
filtering and selection approach:

\begin{enumerate}
  \item Filter configurations to ones with acceptable fairness,
    defined as mean disparate impact between 0.8 and 1.25.
  \item Further filter to ones with nontrivial precision on average,
    i.e., nonzero true positive rate.
  \item Additionally filter configurations to ones with acceptable
    predictive performance, defined as mean F1 (across 5 trials)
    greater than the average of all mean F1 values (average of the
    means over each set of 5 trials) or the median of all mean F1
    values (median of those means), whichever is greater.
  \item Finally, select the mitigator and hyperparameters with maximum
    precision (in case of COMPAS, since true positives should be
    prioritized) or recall (all other datasets, since false negatives
    should be avoided).
\end{enumerate}

Tables~\ref{tab:pre-est-configs} and~\ref{tab:in-est-configs} in the
appendix list the chosen pre-estimator and in-estimator configurations
(the only post-estimator configuration is
\pyplain{CalibratedEqOddsPostprocessing(cost\_constraint="weighted")}).


After the first step is done, the second step comprises the main set
of experiments over the Cartesian product of ensembles and mitigators of
Fig.~\ref{fig:standard_model} plus ensemble hyperparameters. 
For bagging and boosting,
the only ensemble-related hyperparameter varied between configurations
was the number of base estimators used in the ensemble. Values used for bagging configurations included
$\{1,5,10,50,100\}$ and values used for boosting included $\{1,10,50,100,500\}$.

Voting and stacking utilize lists of heterogeneous base estimators as hyperparameters.
In our experiments, these lists contained either $4$ mitigated base estimators or $4$ unmitigated
base estimators (i.e. for a given configuration, either all base estimators were mitigated or none of them were).
When testing in-estimator mitigation with heterogeneous estimators,
all four base estimators are replaced with in-estimator mitigators, specifically hyperparameter-optimized versions of
{PrejudiceRemover, GerryFairClassifier, MetaFairClassifier, and AdversarialDebiasing}.

Lastly, stacking configurations also controlled the value of \textit{passthrough} (whether
dataset features were fed directly to the final estimator) and the mitigation of the final estimator.
Specifically, if \textit{passthrough} was set to \pyplain{True}, either the base estimators
or final estimators could be mitigated, but not both. However, if \textit{passthrough} was set to \pyplain{False},
only the base estimators could be mitigated because the final estimator lacks parameters corresponding
to the dataset features, which in turn are required by mitigation techniques.

Just like the first step, the second step also ran 5~trials of 3-fold
cross validation for each experiment configuration, and we recorded
the same raw and mean-aggregated metrics. 
We used a computing cluster to run these experiments where
each compute node has an Intel Xeon E5-2667 processor @ 3.30GHz.
Every experiment configuration run was allotted 4 cores and 12 GB memory. 

\subsection{Results}\label{sec:results}

To determine whether ensembles help with fairness (and if so, how?), we analyze the metrics from
our Cartesian product evaluation through answering several research questions:

\begin{enumerate}
  \item \textit{Do ensembles help with fairness?}
  \item \textit{Do ensembles help predictive performance when there is mitigation?}
  \item \textit{How does ensemble size affect resource consumption?}
  \item \textit{Can ensemble-level mitigation achieve the same fairness as estimator-level?}
\end{enumerate}

\subsubsection{Result preprocessing steps}

Recall that we use a variety of different datasets. Since these datasets can have vastly different
numbers of examples, feature space sizes, and baseline disparate impact values, learning fair models
is an easier task with some datasets than others. This in turn makes comparing performance
across datasets difficult. We alleviate this problem by applying the following
procedure on a per-dataset basis for each metric of interest: first, we compile
all of the results for each combination of ensemble type and mitigator kind, then
we filter out results with trivial values
for those metrics (corresponding to problems with model fitting).
Subsequently, we map all values to the same region of metric space around the point of optimal fairness
(i.e. for ratio-based metrics where 1 is optimal, we use the reciprocal of a value for downstream calculations 
if the value is larger than 1, and for difference-based metrics where 0 is optimal, we use the absolute value).
Finally, we perform min-max scaling on the mean and standard deviation
of the metric of interest, separately. After doing this for all datasets, we can
group the data by mitigation kind and ensemble type, and average the scaled values
over all datasets for each group to draw meaningful conclusions.

Given a metric \textit{x}, we refer to the metric resulting from
scaling of the mean values of \textit{x} as ``standardized \textit{x}
outcome'' and to the metric resulting from scaling of the values of
\textit{x}'s standard deviation as ``standardized \textit{x} volatility''.
Thus, in the tables and figures that follow, note that the values represented are ones
that have been normalized through this process.

\subsubsection{Do ensembles help with fairness?}

Table~\ref{tab:rq1} shows the results of the normalizing process described
above with disparate
impact as the metric of interest. This table shows that mitigation techniques
almost always improved disparate impact outcomes to some degree, regardless of
whether ensemble learning was used or not.
In general, ensemble learning by itself incurs a slight penalty on disparate impact
compared to the corresponding no-ensemble baseline.
This is an important finding, as it rules out the possibility that solely by using an ensemble
learning technique can one hope to achieve fairer results from an outcome perspective relative to a single estimator.
On the other hand, ensemble learning does generally lower the volatility
of disparate impact. This suggests that ensembles \textit{do} help with fairness,
in particular when mitigation is applied, mainly by improving stability at the
cost of average performance.

\begin{table}[h]
  \begin{tabular}{lcccccccc}
  \toprule
              & \multicolumn{2}{c}{Not Mit.}    & \multicolumn{2}{c}{Pre-}        & \multicolumn{2}{c}{In-}         & \multicolumn{2}{c}{Post-}       \\
              \cmidrule(rl){2-3}                \cmidrule(rl){4-5}                \cmidrule(rl){6-7}                \cmidrule(rl){8-9}
              & SDO            & SDV            & SDO            & SDV            & SDO            & SDV            & SDO            & SDV            \\
  \midrule
  No ensemble & \textbf{0.441} & 0.163          & \textbf{0.741} & 0.350          & \textbf{0.865} & 0.379          & \textbf{0.529} & 0.213          \\
  \addlinespace[0.5em]
  Bagging     & 0.363          & 0.079          & 0.566          & \textbf{0.177} & 0.742          & 0.329          & 0.494          & \textbf{0.066} \\
  Boosting    & 0.407          & 0.065          & 0.723          & 0.394          & 0.803          & \textbf{0.296} & 0.507          & 0.076          \\
  Voting      & 0.322          & \textbf{0.063} & 0.553          & 0.315          & 0.408          & 0.353          & 0.200          & 0.114          \\
  Stacking    & 0.424          & 0.189          & 0.616          & 0.269          & 0.460          & 0.357          & 0.379          & 0.228          \\
  \bottomrule
  \end{tabular}
  \caption{Standardized Disparate impact Outcome (SDO) and Volatility (SDV). Highest SDO and lowest SDV are bolded for each mitigation type.
  Note that SDO and SDV utilize different scales.}
  \label{tab:rq1}
  \end{table}

\subsubsection{Do ensembles help predictive performance when there is mitigation?}

Table~\ref{tab:rq2} shows the results of the normalizing process applied to F1 score. It illustrates that
even with ensemble learning, there is still a trade-off between predictive performance and fairness when bias
is present in the input data. In other words, bias mitigation decreases predictive performance.
Moreover, while the configurations with optimal F1 outcomes are generally
ensembles as opposed to single estimators, these tend to have worse volatility.
Conversely, the configurations that improve stability the most do not have a great
effect on SFO.
Therefore, ensembles \textit{can} help with
predictive performance on average \textit{or} they can help with F1 volatility.

\begin{table}[h]
  \begin{tabular}{lcccccccc}
  \toprule
              & \multicolumn{2}{c}{No Mit.}     & \multicolumn{2}{c}{Pre-}        & \multicolumn{2}{c}{In-}         & \multicolumn{2}{c}{Post-}       \\
              \cmidrule(rl){2-3}                \cmidrule(rl){4-5}                \cmidrule(rl){6-7}                \cmidrule(rl){8-9}
              & SFO            & SFV            & SFO            & SFV            & SFO            & SFV            & SFO            & SFV            \\
  \midrule
  No ensemble & 0.746          & 0.182          & \textbf{0.579} & 0.379          & 0.573          & 0.483          & 0.652          & 0.174          \\
  \addlinespace[0.5em]
  Bagging     & \textbf{0.859} & 0.114          & 0.522          & \textbf{0.146} & 0.567          & \textbf{0.167} & 0.687          & 0.108          \\
  Boosting    & 0.804          & 0.222          & 0.479          & 0.253          & 0.626          & 0.177          & 0.669          & \textbf{0.093} \\
  Voting      & 0.816          & \textbf{0.082} & 0.436          & 0.279          & 0.511          & 0.476          & 0.597          & 0.159          \\
  Stacking    & 0.821          & 0.237          & \textbf{0.579} & 0.497          & \textbf{0.675} & 0.583          & \textbf{0.768} & 0.285          \\
  \bottomrule
  \end{tabular}
  \caption{Standardized F1 outcome (SFO) and volatility (SFV). Highest SFO and lowest SFV are bolded for each mitigation type.}
  \label{tab:rq2}
\end{table}

\subsubsection{How does ensemble size affect resource consumption?}

\begin{figure}
  \begin{subfigure}[b]{0.49\textwidth}
    \centering
    \includegraphics[width=\textwidth]{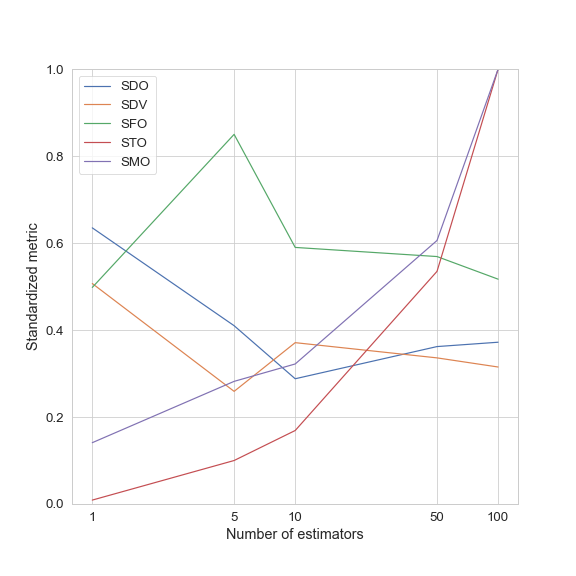}
    \caption{Estimator-level}
  \end{subfigure}
  \begin{subfigure}[b]{0.49\textwidth}
    \centering
    \includegraphics[width=\textwidth]{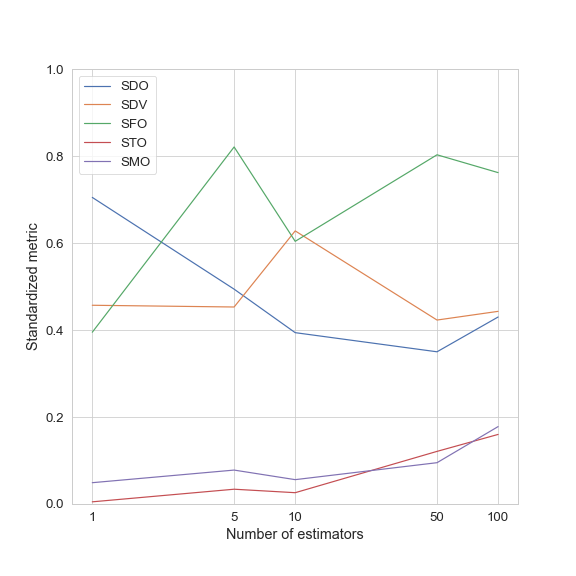}
    \caption{Ensemble-level}
  \end{subfigure}
  \caption{Standardized Time Outcome (STO) and
    Standardized Memory Outcome (SMO) along with previously defined outcome
    and volatility metrics versus number of bagging estimators for ensemble-level and
    estimator-level approaches.}
    \label{fig:resource-consumption}
\end{figure}

Intuitively, we expect that there should be resource consumption differences between
ensemble-level mitigation and estimator-level mitigation when many estimators are used.
Fig.~\ref{fig:resource-consumption} aggregates and displays data in such a way
to analyze this possibility. Specifically, the values plotted correspond to consumed time and memory resources
(in seconds and MB respectively) for pre-estimator-mitigated bagging and different numbers of estimators
mitigated at the ensemble-level versus the estimator-level in order to obtain the associated disparate impact
and F1 results.
As expected, ensemble-level mitigation generally consumes fewer resources in both time and space.
Our final question asks if those savings represent a performance trade-off.

\subsubsection{Can ensemble-level mitigation achieve the same fairness as estimator-level?}

Tables~\ref{tab:rq3-homogeneous} and~\ref{tab:rq3-heterogeneous} show standardized disparate impact outcome and
volatility values per ensemble learning method across all pre-estimator
mitigation techniques and datasets. Table~\ref{tab:rq3-homogeneous}
additionally shows standardized statistical parity difference outcome and volatility
for the homogeneous ensembles. Both tables demonstrate
how group fairness changes as a function of different ensemble
hyperparameters and configurations, especially where
mitigation is performed (at the ensemble level versus at the estimator level).

Given that there are resource trade-offs
associated with performing mitigation at different levels, knowing what levels of resulting mitigation to expect
could be helpful in making a decision regarding how much mitigation to use.
Overall, ensemble-level mitigation and estimator-level mitigation have roughly equivalent
disparate impact outcomes, but estimator-level mitigation tends to have lower volatility.
Moreover, note that the same results hold for statistical parity difference in the homogeneous cases.
Additional results with other group fairness metrics (equal opportunity difference and average odds difference)
can be found in Table~\ref{tab:rq3-homogeneous-supp} located in our appendix.

When configuring stacking, it is best to either not pass the data to the
final estimator or ensure that it is appropriately protected from bias in the data via a mitigation technique.
This is reflected in the extremely poor disparate impact outcome associated with ``Base estimator mitigation; Passthrough; No final mitigation''
in Table~\ref{tab:rq3-heterogeneous}. As stated previously, it is not enough to use an ensemble without proper configuration
(via algorithmic bias mitigators, for instance) and expect less bias.

\begin{table}[]
  \begin{tabular}{lrccccccccc}
  \toprule
  Ensemble type & $n$ & \multicolumn{4}{c}{Estimator-level} & \multicolumn{4}{c}{Ensemble-level} \\
  \cmidrule(rl){3-6}\cmidrule(rl){7-10}
           &                      & SDO            & SDV            & SSO            & SSV            & SDO            & SDV            & SSO            & SSV            \\
  \midrule
  Bagging  & 1                    & \textbf{0.643} & 0.507          & \textbf{0.392} & 0.735          & \textbf{0.713} & 0.457          & 0.541          & 0.693          \\
           & 5                    & 0.414          & 0.259          & 0.632          & 0.287          & 0.499          & 0.453          & 0.532          & 0.617          \\
           & 10                   & 0.294          & 0.371          & 0.604          & 0.298          & 0.394          & 0.628          & 0.531          & 0.522          \\
           & 50                   & 0.363          & 0.336          & 0.489          & \textbf{0.120} & 0.351          & \textbf{0.423} & 0.601          & 0.558          \\
           & 100                  & 0.372          & \textbf{0.315} & 0.549          & 0.179          & 0.430          & 0.443          & \textbf{0.508} & \textbf{0.460} \\
  \addlinespace[0.5em]
  Boosting & 1                    & 0.473          & 0.329          & 0.514          & 0.526          & \textbf{0.730} & 0.430          & \textbf{0.298} & 0.551          \\
           & 10                   & 0.300          & 0.341          & 0.564          & 0.427          & 0.443          & 0.459          & 0.518          & 0.528          \\
           & 50                   & 0.475          & 0.491          & 0.524          & 0.476          & 0.446          & 0.490          & 0.534          & 0.548          \\
           & 100                  & 0.606          & 0.501          & 0.466          & 0.367          & 0.420          & 0.397          & 0.543          & \textbf{0.483} \\
           & 500                  & \textbf{0.672} & \textbf{0.212} & \textbf{0.414} & \textbf{0.214} & 0.393          & \textbf{0.387} & 0.480          & 0.569          \\
  \bottomrule
  \end{tabular}
  \caption{Standardized DI Outcome (SDO) and Volatility (SDV) in addition to
  Standardized Statistical parity difference Outcome (SSO) and Volatility (SSV)
  comparing homogeneous ensembles of various numbers of base estimators ($n$) with
  pre-estimator mitigation where mitigation was performed either at the
  estimator level or ensemble level.
  Highest SDO, lowest SDV, lowest SSO, and lowest SSV are bolded by ensemble type.}
  \label{tab:rq3-homogeneous}
\end{table}

\begin{table}[]
  \begin{tabular}{llcc}
  \toprule
  Ensemble type & Configuration                      & SDO        & SDV    \\
  \midrule
  Voting   & Ensemble-level                     & \textbf{0.462} & 0.615          \\
           & Estimator-level                    & \textbf{0.462} & \textbf{0.308} \\
  \addlinespace[0.5em]
  Stacking & Ensemble-level                     & 0.803          & 0.642          \\
           & Base estimator mitigation; No passthrough                     & \textbf{0.832} & 0.515          \\
           & Base estimator mitigation; Passthrough; No final mitigation   & 0.106          & \textbf{0.229} \\
           & No base estimator mitigation; Passthrough; Only final mitigation & 0.670          & 0.506          \\
  \bottomrule
  \end{tabular}
  \caption{Standardized DI outcome (SDO) and volatility (SDV) comparing heterogeneous ensembles with pre-estimator mitigation.
  Highest SDO and lowest SDV are bolded by ensemble.}
  \label{tab:rq3-heterogeneous}
\end{table}

\section{Guidance for Method Selection}\label{sec:guidance}

What we have summarized thus far are the results of many experiments with various
data and model configurations. One might ask ``\emph{given these results, what
are the best configurations for future experiments?}''

\begin{figure}
\centerline{\includegraphics[width=\textwidth,height=.93\textheight,keepaspectratio]{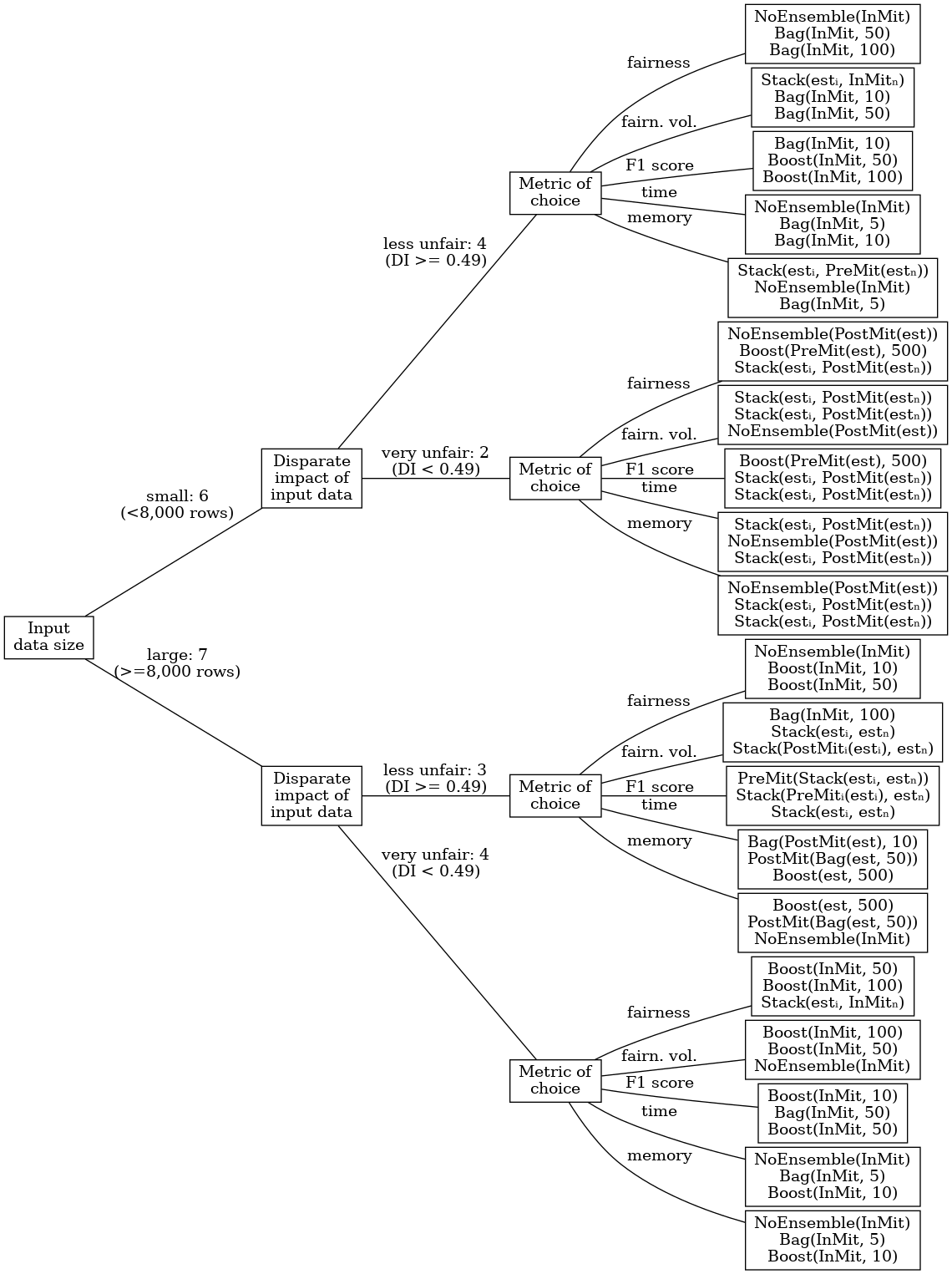}}
\caption{Breakdown of optimal ensembles with respect to metric of choice and dataset configuration.
Edges connecting leaf nodes correspond to Outcome and Volatility metrics described in the previous section.
Each leaf node lists model configurations using the notation
introduced in Fig.~\ref{fig:standard_model} and
Fig.~\ref{lst:lale-mitigator-ensemble-example}.}
\label{fig:tree}
\end{figure}

We attempt to answer this question with Fig.~\ref{fig:tree}, which
displays the best results from our experiments for particular metrics and data setups.
Note that this approach is largely driven by the outcomes of our experiments
with minimal hand-tuning and qualitative analysis to create the final tree.
Specifically, we perform the following steps:

\begin{enumerate}
    \item Organize all results by dataset and create Outcome and Volatility metrics per dataset (as described in the previous section).
    \item Filter results for each dataset to ones that occur in the top 33\% of results for both Standardized Disparate Impact Outcome and Standardized F1 Outcome.
    \item Place each result into one of four groups, or quadrants, based on responses to binary questions related to the corresponding dataset.
    \subitem Is the dataset ``large''? (Yes or No)
    \subitem Is the dataset ``very unfair''? (Yes or No)\\
    Defining ``large'' as containing more than 8,000 rows and ``very unfair'' as having baseline disparate impact under 0.49 led to roughly even
    divisions of results. The Cartesian product of these responses defines the quadrants (i.e. large and unfair, small and fair, etc.).
    \item Average each metric in each quadrant while grouping by model configuration.
    \item Report the top 3 model configurations for each metric in each quadrant.
\end{enumerate}

The notation used in Fig.~\ref{fig:tree} is similar to that of
Fig.~\ref{fig:standard_model}, which in turn resembles Lale syntax.
One key difference is that there is no corresponding syntax in the
previous figure to emphasize the absence of an ensemble; here we use
\texttt{NoEnsemble(...)} to represent these cases.

We hope that such a quantitative analysis of our experimental data and representation
of our results can help practitioners determine best configurations for future experiments.
For instance, our figure suggests using boosting with a large number of in-estimator mitigators
to optimize for fairness on a large and unfair dataset.
Alternatively, it suggests post-estimator mitigation with stacking for a small unfair dataset,
to optimize for a variety of different metrics.

As noted in other parts of this paper, ``best configuration'' and even ``best results'' are highly
dependent on context. The branching and myriad of different settings displayed in this figure additionally
highlight this fact.
While this figure provides guidance for future model-building
experiments and deployments related to algorithmic fairness,
it is intended to merely augment but not replace sound human judgment.

\section{Conclusion}\label{sec:conclusion}
In summary, we have detailed the results of our empirical study
that utilizes the modularity provided by our open-source library
to test various configurations of ensemble learning and mitigation techniques
across thirteen datasets. We find that some configurations work better in certain
situations and yield more stable fairness metrics than others, but regardless of context, fairness and ensemble hyperparameters
must be set properly in order to obtain beneficial results.

We have distilled our findings in the form of a tree in Fig.~\ref{fig:tree} that suggests various
promising models depending on dataset and metric characteristics.
Going forward, we hope that
future practitioners can reproduce our experimental results via our library
and obtain beneficial results in new settings via our guidance diagram.

\bibliography{bibfile}
\bibliographystyle{ACM-Reference-Format}

\appendix
\section{Supplemental Material}\label{sec:supplemental}

\begin{table}[h]
    \begin{tabular}{lrccccccccc}
    \toprule
    Ensemble type & $n$ & \multicolumn{4}{c}{Estimator-level} & \multicolumn{4}{c}{Ensemble-level} \\
    \cmidrule(rl){3-6}\cmidrule(rl){7-10}
             &                      & SAO            & SAV            & SEO            & SEV            & SAO            & SAV            & SEO            & SEV            \\
    \midrule
    Bagging  & 1                    & \textbf{0.487} & 0.763          & \textbf{0.383} & 0.746          & 0.466          & 0.718          & 0.463          & 0.726          \\
             & 5                    & 0.614          & 0.454          & 0.513          & 0.336          & \textbf{0.408} & 0.683          & \textbf{0.384} & 0.595          \\
             & 10                   & 0.503          & 0.305          & 0.442          & 0.313          & 0.500          & 0.570          & 0.500          & 0.456          \\
             & 50                   & 0.555          & \textbf{0.122} & 0.458          & \textbf{0.142} & 0.482          & 0.538          & 0.388          & 0.448          \\
             & 100                  & 0.598          & 0.232          & 0.490          & 0.182          & 0.500          & \textbf{0.459} & 0.616          & \textbf{0.410} \\
    \addlinespace[0.5em]
    Boosting & 1                    & \textbf{0.284} & 0.555          & 0.395          & 0.459          & \textbf{0.346} & 0.605          & 0.457          & 0.708          \\
             & 10                   & 0.538          & 0.325          & 0.550          & 0.409          & 0.546          & 0.571          & 0.483          & 0.409          \\
             & 50                   & 0.726          & 0.378          & 0.554          & 0.465          & 0.418          & 0.505          & \textbf{0.416} & 0.577          \\
             & 100                  & 0.575          & 0.279          & \textbf{0.363} & \textbf{0.358} & 0.586          & \textbf{0.484} & 0.521          & \textbf{0.504} \\
             & 500                  & 0.669          & \textbf{0.237} & 0.536          & 0.388          & 0.505          & 0.489          & 0.489          & 0.613          \\
    \bottomrule
    \end{tabular}
    \caption{Standardized Average odds difference Outcome (SAO) and Volatility (SAV) in addition to
    Standaridzed Equal opportunity difference Outcome (SEO) and Volatility (SEV)
    comparing homogeneous ensembles of various numbers of base estimators ($n$) with
    pre-estimator mitigation where mitigation was performed either at the
    estimator level or ensemble level.
    Lowest SAO, lowest SAV, lowest SEO, and lowest SEV are bolded by ensemble type.}
    \label{tab:rq3-homogeneous-supp}
  \end{table}

\begin{table}[h]
    \begin{tabular}{lll}
    \toprule
    Dataset        & Mitigator              & Hyperparameters            \\
    \midrule
    COMPAS Violent & DisparateImpactRemover & 1                          \\
    Credit-g       & LFR                    & k=5, Ax=0.01, Ay=10, Az=5  \\
    COMPAS         & DisparateImpactRemover & 0.4                        \\
    Ricci          & LFR                    & k=5, Ax=0.01, Ay=5, Az=10  \\
    TAE            & LFR                    & k=5, Ax=0.01, Ay=50, Az=5  \\
    Titanic        & DisparateImpactRemover & 0.8                        \\
    SpeedDating    & DisparateImpactRemover & 0.2                        \\
    Bank           & DisparateImpactRemover & 0.2                        \\
    MEPS 19        & LFR                    & k=5, Ax0.01, Ay=1, Az=10   \\
    MEPS 20        & LFR                    & k=5, Ax=0.01, Ay=1, Az=10  \\
    Nursery        & LFR                    & k=20, Ax=0.01, Ay=1, Az=10 \\
    MEPS 21        & LFR                    & k=5, Ax=0.01, Ay=1, Az=10  \\
    Adult          & LFR                    & k=5, Ax=0.01, Ay=1, Az=10  \\
    \bottomrule
    \end{tabular}
    \caption{Optimal pre-estimator mitigator configurations (with corresponding hyperparameters) per dataset.
    Hyperparameter names are not provided if the mitigation technique only accepts one.
    If a hyperparameter is not listed in the rightmost column, the configuration utilizes the default value.}
    \label{tab:pre-est-configs}
\end{table}

\begin{table}[h]
    \begin{tabular}{lll}
    \toprule
    Dataset        & Mitigator            & Hyperparameters                    \\
    \midrule
    COMPAS Violent & MetaFairClassifier   & 0.5                                \\
    Credit-g       & AdversarialDebiasing & classifier\_num\_hidden\_units=10  \\
    COMPAS         & MetaFairClassifier   & 0.5                                \\
    Ricci          & MetaFairClassifier   & 0.8                                \\
    TAE            & MetaFairClassifier   & 0.8                                \\
    Titanic        & MetaFairClassifier   & 1                                  \\
    SpeedDating    & MetaFairClassifier   & 0.9                                \\
    Bank           & PrejudiceRemover     & 100                                \\
    MEPS 19        & PrejudiceRemover     & 1000                               \\
    MEPS 20        & AdversarialDebiasing & classifier\_num\_hidden\_units=500 \\
    Nursery        & MetaFairClassifier   & 0.5                                \\
    MEPS 21        & AdversarialDebiasing & classifier\_num\_hidden\_units=500 \\
    Adult          & PrejudiceRemover     & 1000                               \\
    \bottomrule
    \end{tabular}
    \caption{Optimal in-estimator mitigator configurations (with corresponding hyperparameters) per dataset.
    Hyperparameter names are not provided if the mitigation technique only accepts one.
    If a hyperparameter is not listed in the rightmost column, the configuration utilizes the default value.}
    \label{tab:in-est-configs}
\end{table}


\end{document}